\documentclass[letterpaper]{article} 
\usepackage[preprint]{aaai2027}  
\usepackage[hyphens]{url}  
\usepackage{graphicx} 
\usepackage{natbib}  
\usepackage{caption} 
\usepackage{algorithm}
\usepackage{algorithmic}

\usepackage{newfloat}
\usepackage{listings}
\DeclareCaptionStyle{ruled}{labelfont=normalfont,labelsep=colon,strut=off} 
\floatstyle{ruled}
\newfloat{listing}{tb}{lst}{}
\floatname{listing}{Listing}

\usepackage{booktabs}
\usepackage{amssymb}
\usepackage{multirow}
\usepackage{tabularx}

\title{TempoGround: State-Aware Streaming Visual Grounding with Vision-Language Models}
\author{
    Leqian Ding\textsuperscript{\rm 1}\thanks{This work was completed during an internship at EngineAI.},
    Junning Qiu\textsuperscript{\rm 2}\corresponding,
    Manwen Yang\textsuperscript{\rm 1},
    Yu Guo\textsuperscript{\rm 1}\corresponding,
    Fei Wang\textsuperscript{\rm 1}
}
\affiliations{
    \textsuperscript{\rm 1}Xi'an Jiaotong University, Xi'an, China\\
    \textsuperscript{\rm 2}EngineAI, Shenzhen, China
}

\begin{document}

\maketitle

\begin{abstract}

Visual grounding maps language referents to spatial targets and is central to open-vocabulary perception with vision-language models. Existing methods have made substantial progress on single-frame and video-based visual grounding, yet under streaming inputs they still suffer from identity drift, cross-frame inconsistency, and fragile localization under partial occlusion. To address these issues, we present \textbf{TempoGround}, a VLM-native framework that detects cross-frame object correspondence and explicitly models object presence states, thereby enabling accurate and consistent visual grounding under streaming inputs. The key is a curriculum prediction mechanism guided by state-aware cross-frame correspondence: TempoGround resolves 2D instance association, predicts whether each object newly enters, continues in, or leaves the view, decodes the 2D box, and then lifts it to a camera-frame 3D box. As token-level supervision alone cannot capture the geometric objectives of streaming grounding, we further introduce Streaming Grounding Reinforcement (SGR), which optimizes TempoGround with verifiable Grounding, Identity, and Consistency rewards, jointly reinforcing persistent localization and temporally consistent predictions. We carefully design a three-stage training strategy and train TempoGround on large-scale data. We evaluate visual grounding under causally streaming inputs on multiple challenging benchmarks: TempoGround improves F1$_{2D}@0.5$ and F1$_{2D}@0.95$ by 4.4 and 0.5 on average, and F1$_{3D}@0.25$ and AP$_{3D}$ by 6.2 and 7.5, respectively. These results demonstrate that TempoGround provides a practical foundation for visual grounding under streaming inputs.

\end{abstract}


\section{Introduction}

\begin{figure*}[t]
\centering
\includegraphics[width=\textwidth]{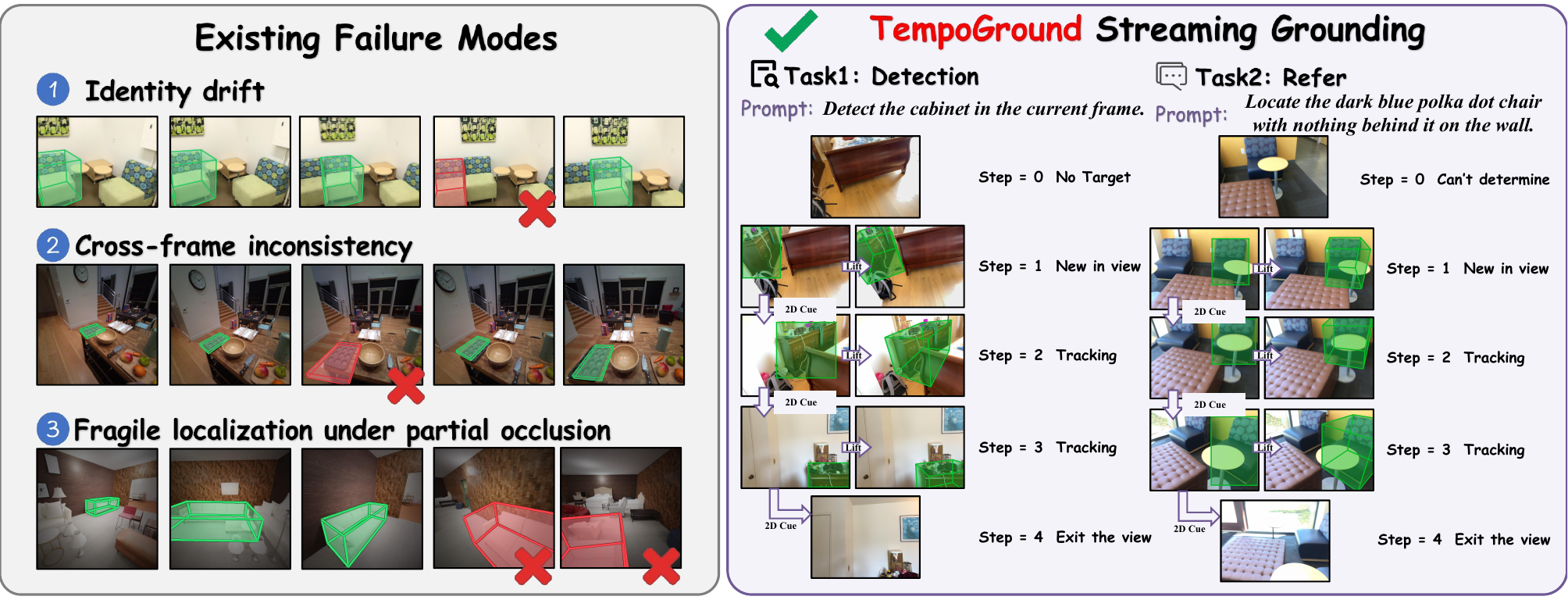}
\caption{\textbf{TempoGround for streaming visual grounding.} Under streaming inputs, visual grounding commonly exhibits three characteristic failure modes (left). By exploiting cross-frame correspondence for state-aware curriculum prediction, TempoGround yields more stable camera-frame 2D/3D localization under streaming inputs (right) for both detection and referring grounding.}
\label{fig:teaser}
\end{figure*}

Vision-language models (VLMs) have made visual grounding a practical route to open-vocabulary perception: given a free-form language query, the model localizes the referred objects as spatially actionable targets, which is important for embodied perception and related downstream control. Most existing VLM grounders operate on single-frame or video-based inputs. In practical deployments, however, sensory observations are delivered as a causal stream. We define streaming visual grounding as the setting in which a language query is given, images arrive incrementally, and the model must continuously track the matching objects in view and produce their 2D/3D bounding boxes. As shown in Figure~\ref{fig:teaser}, streaming visual grounding commonly exhibits three characteristic failure modes. Identity drift confuses same-category instances across time. Cross-frame inconsistency yields temporally unstable boxes across adjacent frames. Fragile localization under partial occlusion arises when intermittent visibility causes missed or degraded detections that cannot be corrected without causal cross-frame reasoning.Consequently, visual grounding under streaming inputs remains challenging.

Recent 3D scene understanding methods enhance object-level perception and grounding by injecting globally aligned geometric information. These approaches typically perform object grounding in a world or scene coordinate system and therefore introduce a global coordinate dependency. For object-level perception under ego-centric viewpoints, as in embodied manipulation and related scenarios, world-frame predictions are not directly applicable in the current camera view and require an additional alignment step that introduces additional alignment error. We therefore aim for a VLM-native method for streaming visual grounding that consumes causal image streams and predicts in the current camera frame.

To address this limitation, we present TempoGround with a state-aware streaming curriculum prediction mechanism that leverages cross-frame correspondence for object state prediction and spatial localization. At each timestep, under language guidance, TempoGround first associates instances across adjacent frames by matching a lightweight 2D cue from the previous prediction against corresponding regions in the current image, thereby maintaining consistent object identity. Based on this correspondence, TempoGround then estimates whether each object newly enters, persists in, or exits the field of view. Conditioned on these correspondence and presence cues, TempoGround better captures cross-frame object states and spatial localization. As reliable 3D predictions benefit from stable 2D evidence~\cite{man2026locateanything3d}, TempoGround first predicts a 2D box for each instance and then lifts it to a camera-frame 3D box. Taken together, TempoGround follows a curriculum prediction chain that resolves 2D correspondence, estimates object presence, decodes the 2D box, and lifts the prediction to 3D. This design allows TempoGround to exploit cross-frame instance correspondence and spatial position changes under streaming inputs, supporting more robust and accurate tracking.

As a VLM-native approach to streaming object grounding, TempoGround is built on a standard VLM decoder, produces outputs in a standard autoregressive manner, and is trained with standard supervised fine-tuning on carefully constructed large-scale data. However, supervised fine-tuning alone remains insufficient, because token-level supervision treats bounding box prediction as an autoregressive token sequence and therefore does not directly model the geometric attributes of visual bounding boxes, nor does it provide direct box-level supervision under streaming inputs. To meet the consistency requirements of streaming grounding, we introduce Streaming Grounding Reinforcement (SGR), an RLVR stage that optimizes TempoGround with verifiable Grounding, Identity, and Consistency rewards. These rewards target localization quality along the stream, correct enter and exit decisions, and stable cross-frame predictions.

Building on the above considerations, we construct a three-stage training recipe that progresses from single-frame geometric competence to streaming supervision and finally to sequence-level reinforcement. Stage 1 applies supervised fine-tuning on \textbf{26M} samples to strengthen large-scale 2D detection and 2D-to-3D lifting, establishing the spatial foundation for subsequent learning. Building on this competence, Stage 2 continues with supervised fine-tuning on \textbf{5.1M} samples to establish curriculum prediction, enabling cross-frame correspondence to support presence estimation and consistent localization. Stage 3 then refines these streaming capabilities with SGR under verifiable rewards, aligning training with the geometric objectives that token-level supervision alone cannot capture.

Our contributions are as follows:
\begin{itemize}
\item We propose \textbf{TempoGround}, a VLM-native framework that enables accurate and consistent streaming visual grounding through \textbf{curriculum prediction} guided by state-aware cross-frame correspondence, from 2D association and presence estimation to camera-frame 3D grounding.
\item We introduce \textbf{Streaming Grounding Reinforcement (SGR)}, an RLVR method that optimizes TempoGround with verifiable Grounding, Identity, and Consistency rewards to align training with the geometric objectives of streaming grounding.
\item Extensive experiments on multiple challenging benchmarks under causally streaming inputs show that TempoGround improves F1$_{2D}$ at IoU thresholds 0.5 and 0.95 by 4.4 and 0.5 on average, and improves F1$_{3D}$ at IoU 0.25 and AP$_{3D}$ by 6.2 and 7.5, providing a practical foundation for streaming visual grounding.
\end{itemize}
\section{Related Works}

\subsection{VLM-based Visual Perception}

Vision-language models have been increasingly applied to spatial understanding in 3D environments. Reconstructed geometry or multi-view video can be encoded into LLMs for relational and layout-level dialogue~\cite{hong20233d}, coupled with embodied reasoning and control~\cite{huang2023embodied}, or unified with grounding, captioning, and question answering under shared instruction following~\cite{huang2024chat,zheng2025video}. Under camera motion, explicit geometry is further injected through position-aware visual tokens~\cite{zhu2024llava}, globally consistent geometric fusion~\cite{zheng2026learning}, or reconstructive 3D tokens for spatial reasoning~\cite{fan2025vlm}, while first-person 3D sensing resources continue to expand~\cite{wang2024embodiedscan}. These methods improve scene-level semantics and spatial reasoning over offline reconstructions and short clips, yet their supervision and inference remain largely scan- or clip-centric, and object predictions are often formulated in a world or scene coordinate system. TempoGround instead targets object-level localization in the current camera frame under causally arriving image streams.

\subsection{Visual Object Grounding}

Visual object grounding associates language queries with spatial targets. Open-vocabulary detectors condition localization on category names or referring expressions~\cite{liu2024grounding,chen2020scanrefer}, while multimodal LLMs cast boxes as autoregressive tokens~\cite{peng2024grounding,bai2025qwen25vl,chen2023shikra}. Subsequent work strengthens this generative formulation through pixel-aligned grounding~\cite{rasheed2024glamm}, high-throughput generative detection~\cite{jiang2025detect,wang2026locateanything}, and staged 2D-to-3D prediction in the camera frame~\cite{man2026locateanything3d}. Reinforcement with verifiable rewards further aligns generated geometry with evaluation metrics~\cite{linghu20263d}. These advances mainly address single-frame or offline clip settings, where each observation can be treated largely independently or with access to the full sequence. Streaming visual grounding additionally requires identity-stable and cross-frame-consistent box updates as images arrive causally. TempoGround addresses this setting.
\section{Method}
\label{sec:method}

\begin{figure*}[t]
\centering
\includegraphics[width=\textwidth]{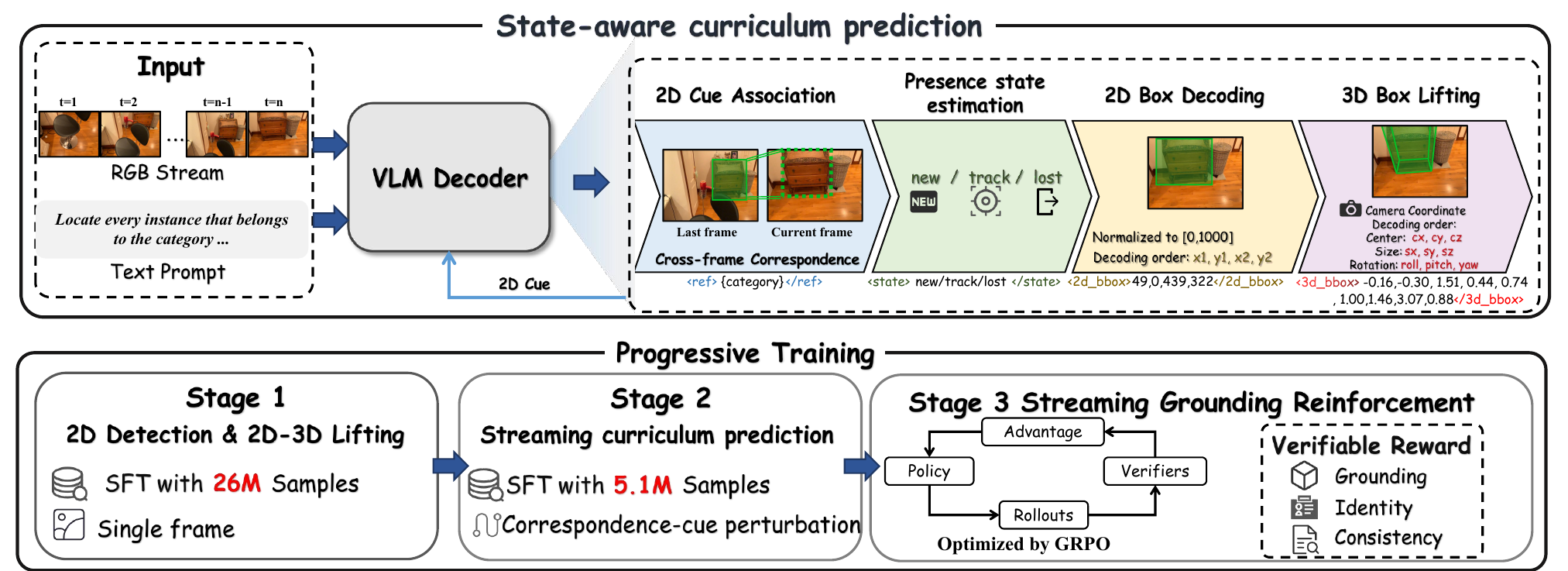}
\caption{\textbf{TempoGround overview.}
Top: state-aware curriculum prediction guided by cross-frame correspondence, consuming adjacent dual-image frames with the language query and previous 2D cue, from presence-aware 2D localization to camera-frame 3D lifting.
Bottom: progressive three-stage training from single-frame SFT to streaming curriculum prediction and Streaming Grounding Reinforcement (SGR).}
\label{fig:framework}
\end{figure*}

We present TempoGround, a VLM-native framework for streaming visual grounding.
Figure~\ref{fig:framework} overviews the state-aware curriculum prediction interface and the progressive three-stage training recipe.
Section~\ref{sec:formulation} formalizes the streaming visual grounding task and the geometry conventions used throughout.
Section~\ref{sec:curriculum} describes curriculum prediction guided by state-aware cross-frame correspondence.
Section~\ref{sec:sgr} introduces Streaming Grounding Reinforcement (SGR) with verifiable Grounding, Identity, and Consistency rewards.
Section~\ref{sec:training} presents the progressive three-stage training recipe.
Section~\ref{sec:data} summarizes data curation for single-frame and streaming supervision.
Additional implementation details are deferred to the appendix.

\subsection{Problem Formulation}
\label{sec:formulation}

\paragraph{Streaming visual grounding.}
Let $q$ denote a free-form language query, either a category name (detection) or a referring expression (refer).
Let $I_{1:T}=(I_1,\ldots,I_T)$ be a causally arriving RGB stream, where $I_t\in\mathrm{R}^{H\times W\times 3}$ is the observation at timestep $t$.
At each timestep $t$, the model must localize all objects in view that match $q$ and output their $2$D boxes and camera-frame $3$D boxes.
Formally, the prediction at timestep $t$ is a (possibly empty) set
\begin{equation}
\label{eq:pred-set}
\mathcal{Y}_t=\left\{y_t^{(k)}\right\}_{k=1}^{N_t},
\qquad
y_t^{(k)}=\left(\mathbf{b}_{t}^{2\mathrm{D},(k)},\,\mathbf{b}_{t}^{3\mathrm{D},(k)}\right),
\end{equation}
where $N_t$ is the number of predicted instances,
$\mathbf{b}_{t}^{2\mathrm{D},(k)}\in\mathrm{R}^{4}$ is an axis-aligned $2$D box, and
$\mathbf{b}_{t}^{3\mathrm{D},(k)}\in\mathrm{R}^{9}$ is a camera-frame $3$D box.
The streaming setting is causal: the prediction at timestep $t$ may depend only on $q$, observations $I_{1:t}$, and previous predictions $\mathcal{Y}_{1:t-1}$.

\paragraph{Geometry conventions.}
All $2$D boxes are represented in normalized image coordinates
$\mathbf{b}^{2\mathrm{D}}=(x_1,y_1,x_2,y_2)$ with each coordinate quantized to integers in $[0,1000]$.
All $3$D boxes are expressed in the current camera frame as
\begin{equation}
\label{eq:bbox3d}
\mathbf{b}^{3\mathrm{D}}=(c_x,c_y,c_z,s_x,s_y,s_z,r,p,y),
\end{equation}
where $(c_x,c_y,c_z)$ is the box center in meters,
$(s_x,s_y,s_z)$ is the size in meters, and
$(r,p,y)$ denotes roll, pitch, and yaw in radians.

\subsection{Curriculum Prediction with State-Aware Cross-Frame Correspondence}
\label{sec:curriculum}

TempoGround is built on a standard autoregressive VLM decoder.
At timestep $t$, under the guidance of $q$, the decoder emits a structured token sequence that realizes the following curriculum:
\textit{2D correspondence $\rightarrow$ presence estimation $\rightarrow$ 2D box decoding $\rightarrow$ camera-frame 3D lifting}.

\paragraph{Cross-frame correspondence.}
Let $\hat{\mathcal{Y}}_{t-1}$ denote the model prediction at the previous timestep (empty when $t=1$).
At each timestep $t>1$, TempoGround takes both the previous image $I_{t-1}$ and the current image $I_t$ as visual inputs.
We construct a lightweight correspondence cue
\begin{equation}
\label{eq:cue}
\mathbf{c}_{t-1}
=
\left\{\left(\ell_{t-1}^{(k)},\,\mathbf{b}_{t-1}^{2\mathrm{D},(k)}\right)\right\}_{k},
\end{equation}
which retains only labels and $2$D boxes from $\hat{\mathcal{Y}}_{t-1}$ and discards $3$D boxes.
Camera-frame $3$D boxes are coupled to the viewpoint at the previous timestep; carrying them forward would inject stale geometric bias into the current frame.
The $2$D cue instead provides a view-stable association signal, after which TempoGround re-estimates $3$D in the current camera frame.
Conditioned on the adjacent image pair $(I_{t-1},I_t)$, language query $q$, and correspondence cue $\mathbf{c}_{t-1}$, TempoGround associates instances across frames by matching $\mathbf{c}_{t-1}$ against corresponding regions in $I_t$.

\paragraph{Presence state estimation.}
To strengthen cross-frame instance association, TempoGround predicts a discrete presence state for each associated instance.
We use the presence vocabulary
\begin{equation}
\label{eq:presence}
\mathcal{S}=\{\texttt{new},\,\texttt{track},\,\texttt{lost}\},
\end{equation}
where $\texttt{new}$, $\texttt{track}$, and $\texttt{lost}$ respectively indicate that an object newly enters the field of view, persists in the view, or exits the view.
A predicted state $s_t\in\mathcal{S}$ determines whether the corresponding instance contributes a box to $\mathcal{Y}_t$.
If a previously lost query-matching object re-enters the view, TempoGround treats it as $\texttt{new}$ again: presence is defined with respect to contiguous visibility under language guidance, rather than as a persistent identity memory across exits.
Supervising these states is critical for suppressing identity drift among currently visible instances and for preventing continued box predictions after objects leave the view.

\paragraph{From $2$D boxes to camera-frame $3$D.}
After correspondence and presence are resolved, TempoGround decodes a $2$D box for each visible instance and then lifts it to a camera-frame $3$D box of the form $\mathbf{b}^{3\mathrm{D}}=(c_x,c_y,c_z,s_x,s_y,s_z,r,p,y)$.
Predicting $2$D first and then lifting to $3$D lets the $2$D box serve as an intermediate spatial anchor that stabilizes monocular camera-frame $3$D estimation.
Multi-instance responses are serialized in a canonical order.
The resulting autoregressive target interleaves presence, $2$D, and $3$D tokens in the same order used at inference.

\subsection{Streaming Grounding Reinforcement}
\label{sec:sgr}

Token-level supervised fine-tuning trains the decoder to imitate discrete coordinate tokens.
This objective cannot directly model continuous box geometry, nor does it provide trajectory-level supervision when each step depends on previous model outputs.
We therefore introduce Streaming Grounding Reinforcement (SGR), an RLVR stage that optimizes TempoGround with verifiable rewards on streamed rollouts.

\paragraph{Objective.}
Given a streaming prompt $x=(q,I_{1:T})$, the policy $\pi_{\theta}$ samples a trajectory $\tau=\mathcal{Y}_{1:T}$.
A deterministic verifier returns a scalar reward $R(\tau)$.
SGR maximizes the expected reward
\begin{equation}
\label{eq:sgr-obj}
J(\theta)
=
\mathrm{E}_{x\sim\mathcal{D},\,\tau\sim\pi_{\theta}(\cdot\mid x)}
\left[R(\tau)\right].
\end{equation}
We optimize this objective with Group Relative Policy Optimization (GRPO)~\cite{shao2024deepseekmath}.
For each prompt, GRPO samples a group $\{\tau_i\}_{i=1}^{G}$ from the old policy and computes the group-normalized advantage
\begin{equation}
\label{eq:grpo-adv}
A_i
=
\frac{R(\tau_i)-\mathrm{mean}(\{R(\tau_j)\}_{j=1}^{G})}
{\mathrm{std}(\{R(\tau_j)\}_{j=1}^{G})+\epsilon},
\end{equation}
where $\epsilon>0$ is a small constant for numerical stability.
During SGR rollouts, the correspondence cue $\mathbf{c}_{t-1}$ is taken from the model's own previous prediction, so optimization reflects closed-loop streaming behavior.

\begin{figure}[t]
\centering
\includegraphics[width=\linewidth]{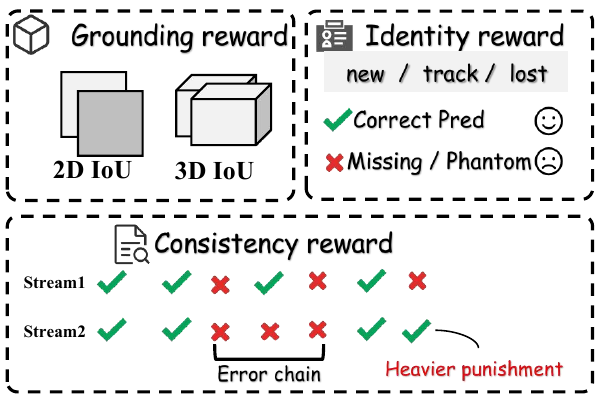}
\caption{Verifiable rewards in SGR.}
\label{fig:sgr-rewards}
\end{figure}

\paragraph{Verifiable rewards.}
SGR assigns each closed-loop trajectory a scalar reward
\begin{equation}
\label{eq:reward}
R(\tau)
=
R_{\mathrm{g}}(\tau)
+
R_{\mathrm{id}}(\tau)
+
R_{\mathrm{con}}(\tau)
-
f_{\mathrm{fmt}}(\tau),
\end{equation}
where $f_{\mathrm{fmt}}$ penalizes unparsable outputs.
As illustrated in Figure~\ref{fig:sgr-rewards}, The three main terms are:
\begin{itemize}
\item \textbf{Grounding} $R_{\mathrm{g}}$:
encourages accurate and persistent $2$D/$3$D boxes for query-matching objects throughout the stream;
\item \textbf{Identity} $R_{\mathrm{id}}$:
encourages correct enter and exit behavior, so objects are detected when they appear and are not predicted after they leave;
\item \textbf{Consistency} $R_{\mathrm{con}}$:
discourages long streaks of failures that propagate through self-generated correspondence cues.
\end{itemize}
Computation details are provided in the appendix.

\subsection{Progressive Three-Stage Training}
\label{sec:training}

We train TempoGround with a progressive three-stage curriculum that moves from single-frame geometric competence to streaming supervision and finally to sequence-level reinforcement.

\paragraph{Stage 1: Single-frame $2$D detection and $2$D-to-$3$D lifting.}
Stage~1 applies supervised fine-tuning on about $26$M single-frame samples covering large-scale $2$D detection/referring and camera-frame $3$D detection/referring.
The model learns open-vocabulary localization and the $2$D-then-$3$D decoding order that underlies later curriculum prediction.

\paragraph{Stage 2: Streaming curriculum prediction.}
Stage~2 continues supervised fine-tuning on about $5.1$M samples.
Each sample provides an adjacent image pair $(I_{t-1},I_t)$, a language query, and a correspondence cue derived from the previous frame, and supervises presence states together with $2$D/$3$D boxes.
This stage installs state-aware cross-frame correspondence so that presence estimation and consistent localization become learnable under temporal cues.

\paragraph{Stage 3: SGR.}
Stage~3 refines the Stage~2 policy with SGR on streamed trajectories under the objective and rewards introduced above.
This stage aligns training with geometric and consistency objectives that token-level imitation cannot capture.

\subsection{Data Curation}
\label{sec:data}

\begin{table*}[!t]
\centering
\small
\setlength{\tabcolsep}{3pt}
\caption{Training sources for TempoGround.
Stage~1 uses $15.6$M $2$D samples and $10.5$M camera-frame $2$D+$3$D samples.
Stage~2 uses $5.1$M samples, from which Stage~3 draws a subset.}
\label{tab:data-sources}
\begin{tabularx}{\textwidth}{@{}c c >{\raggedright\arraybackslash}X c@{}}
\toprule
\textbf{Stage} & \textbf{Modality} & \multicolumn{1}{c}{\textbf{Sources}} & \textbf{Samples} \\
\midrule
\multirow{2}{*}[-5ex]{1}
& $2$D
& \textbf{Objects365}~\cite{shao2019objects365}, \textbf{V3Det}~\cite{wang2023v3det}, \textbf{COCO}~\cite{lin2014microsoft}, \textbf{LVIS}~\cite{gupta2019lvis}, \textbf{EgoObjects}~\cite{zhu2023egoobjects}, \textbf{BDD100K}~\cite{yu2020bdd100k}, \textbf{nuImages}~\cite{caesar2020nuscenes}, \textbf{PACO}~\cite{ramanathan2023paco}, \textbf{RefCOCO/+/g}~\cite{kazemzadeh2014referitgame,yu2016modeling}, \textbf{RefL4}~\cite{chen2025revisiting}, \textbf{Flickr30k Entities}~\cite{plummer2015flickr30k}
& $15.6$M \\
& $2$D+$3$D
& \textbf{ScanNet++}~\cite{yeshwanth2023scannet++}, \textbf{ScanNet}~\cite{dai2017scannet}, \textbf{ADT}~\cite{pan2023aria}, \textbf{ASE}~\cite{avetisyan2024scenescript}, \textbf{ARKitScenes}~\cite{baruch2021arkitscenes}, \textbf{Hypersim}~\cite{roberts2021hypersim}, \textbf{Objectron}~\cite{ahmadyan2021objectron}, \textbf{SUN-RGBD}~\cite{song2015sun}, \textbf{KITTI}~\cite{geiger2012we}, \textbf{nuScenes}
& $10.5$M \\
\midrule
2--3 & Stream
& \textbf{ARKitScenes}, \textbf{ASE}, \textbf{ADT}, \textbf{ScanNet++}, \textbf{ScanNet}
& $5.1$M \\
\bottomrule
\end{tabularx}
\end{table*}

We curate a camera-centric corpus that supports both single-frame Stage~1 supervision and streaming Stages~2--3.
Heterogeneous public sources are converted into a unified JSONL schema and packaged as VLM conversations whose response order matches the curriculum above.
Detailed filtering thresholds, packaging templates, and per-source statistics are provided in the appendix.

\paragraph{Sources and scale.}
Table~\ref{tab:data-sources} summarizes training sources.
Stage~1 mixes large-scale $2$D corpora with camera-frame $2$D+$3$D corpora, totaling about $26$M samples.
Stages~2--3 are built from multi-view RGB-D sequences with stable object identities.
Stage~2 uses $5.1$M streaming samples, from which Stage~3 draws a subset for SGR.

\paragraph{Unified geometry interface.}
Every sample is stored in a shared JSONL format with image pointers, language queries, instance lists, and camera intrinsics.
Box normalization and geometry conventions follow Section~\ref{sec:formulation}.
This unification removes dataset-specific heads and keeps Stage~1 $3$D supervision consistent with streaming camera-frame prediction.

\paragraph{Quality control.}
Aggregating heterogeneous public sources inevitably introduces inconsistent category granularity and unreliable geometry.
We canonicalize ambiguous or overly fine-grained labels, remove geometrically degenerate classes, and discard low-quality boxes through frustum, visibility, and consistency checks.
Calibrated negatives are further included so that absent queries teach rejection rather than hallucination.
On streaming sources, presence-state labels follow per-frame visibility, and clips are sampled with randomized length and stride to diversify temporal contexts.
Concrete filtering thresholds and cleaning pipelines are provided in the appendix.

\paragraph{Correspondence-cue perturbation.}
A natural concern for cue-conditioned streaming is error amplification: once the previous prediction is wrong, under fast ego-motion the previous $2$D box may misalign severely with the current image, and a model that over-relies on $\mathbf{c}_{t-1}$ can drift across subsequent frames.
To prevent such brittle dependence, Stage~2 constructs $\mathbf{c}_{t-1}$ from a mixture of clean ground-truth cues, Stage~1 rollouts, and controlled synthetic perturbations, while keeping supervised targets unchanged.
Table~\ref{tab:cue-perturbation} lists the non-clean modes, which jointly cover $80\%$ of Stage~2 samples; the remaining $20\%$ use clean previous-frame ground-truth $2$D boxes.
Rollout cues provide realistic closed-loop errors; synthetic modes further cover ego-motion misalignment, missed and spurious detections, stale cues, and stream restarts.
Implementation details of each operator are deferred to the appendix.

\begin{table*}[t]
\centering
\small
\setlength{\tabcolsep}{6pt}
\caption{Non-clean correspondence-cue modes in Stage~2 ($80\%$ of samples; the remaining $20\%$ use clean GT cues).
Each mode corrupts only the input prior $\mathbf{c}_{t-1}$; supervision targets remain unchanged.}
\label{tab:cue-perturbation}
\begin{tabularx}{\textwidth}{@{}
>{\raggedright\arraybackslash}p{0.12\textwidth}
>{\centering\arraybackslash}p{0.12\textwidth}
>{\raggedright\arraybackslash}X@{}}
\toprule
\textbf{Mode} & \textbf{Ratio} & \textbf{Role in training} \\
\midrule
Stage~1 rollout & $25\%$ & Uses frozen Stage~1 predictions as cues to expose realistic closed-loop errors. \\
Jitter & $20\%$ & Shifts prior $2$D boxes to simulate fast ego-motion and cross-frame misalignment. \\
Dropout & $12\%$ & Removes prior instances to teach recovery from previous-step misses. \\
Phantom & $12\%$ & Inserts false boxes or wrong labels to discourage trusting spurious cues. \\
Stale & $7\%$ & Replaces the $t{-}1$ cue with an older cue to handle outdated association. \\
Empty & $4\%$ & Clears the cue to force re-initialization after a stream break. \\
\bottomrule
\end{tabularx}
\end{table*}

\paragraph{View-grounded referring annotation.}
Public datasets such as ScanRefer provide scene-level referring expressions that are not suited to map-free streaming, so we discard them as primary supervision and regenerate view-grounded queries on ego imagery.
Each expression avoids absolute directional phrases tied to the global scene layout and instead relies on target attributes and relations to co-visible anchors in the current view.
Along a stream, the target and its anchors jointly determine the first resolvable timestep: the object is labeled $\texttt{new}$ only when the query becomes uniquely decidable from the observed view, after which presence states follow the target with cross-frame correspondence.
Annotation protocols and uniqueness checks are detailed in the appendix.
\section{Experiments}
\label{sec:experiments}

\subsection{Implementation Details}
\label{sec:impl}

\paragraph{Training details.}
We initialize from Qwen3.5 and fully fine-tune all parameters with AdamW in ms-swift~\cite{zhao2025swift}.
We use bfloat16, FlashAttention, and DeepSpeed ZeRO-3, with weight decay $0.05$ and gradient clipping at $1.0$.
All training runs on $128$ Alibaba Cloud PPUs and takes about $109$ hours.

Stage~1 trains for $1$ epoch with learning rate $4{\times}10^{-5}$, vision learning rate $2{\times}10^{-5}$, sequence length $8192$, packing length $8192$, and global batch size $1024$, taking about $69$ hours.

Stage~2 uses the same schedule for $1$ epoch with sequence length $16384$, packing length $16384$, and global batch size $1024$, taking about $21$ hours.

Stage~3 trains with GRPO for $2$ epochs on about $4.1$K streaming trajectories, with $32$ prompts per step, learning rate $1{\times}10^{-6}$, vision learning rate $5{\times}10^{-7}$, KL coefficient $\beta{=}0.01$, $G{=}4$ rollouts at temperature $1.0$, and up to $15$ streaming turns, taking about $19$ hours.

\paragraph{Evaluation setup.}
To ensure a fair comparison, we evaluate all methods under the same causal streaming protocol.
Single-frame methods take each frame independently; video-based methods take the clip observed so far at timestep $t$ and report only the current-frame prediction.
Each dataset contributes $24$ test sequences with equal detection and referring samples.
We report $\mathrm{F1}_{2\mathrm{D}}@0.5$/$@0.95$ for $2$D and $\mathrm{F1}_{3\mathrm{D}}@0.25$/$\mathrm{AP}_{3\mathrm{D}}$ for $3$D.
These metrics are sufficient for our setting because streaming failures ultimately surface as localization errors: incorrect enter/exit or identity decisions cause misses and spurious boxes, while cross-frame inconsistency yields unstable matches over time.
$\mathrm{F1}$ averages per-frame precision and recall along the stream and therefore reflects whether presence and correspondence remain correct at typical timesteps.
$\mathrm{AP}_{3\mathrm{D}}$ aggregates predictions over the full trajectory and is more sensitive to persistent false positives and fragmented tracks under closed-loop updates.
Together, they evaluate both step-wise grounding quality and stream-level stability without requiring separate state-only scores as primary metrics.

\subsection{Main Results}
\label{sec:main-results}

\begin{table*}[t]
\centering
\small
\setlength{\tabcolsep}{3.5pt}
\caption{$3$D grounding results.
Best results are in \textbf{bold}; second best are \underline{underlined}.}
\label{tab:main-3d}
\begin{tabular*}{\textwidth}{@{\extracolsep{\fill}}lcccccccc@{}}
\toprule
\multirow{2}{*}[-0.5ex]{\textbf{Method}}
& \multicolumn{2}{c}{\textbf{ARKitScenes}}
& \multicolumn{2}{c}{\textbf{ScanNet}}
& \multicolumn{2}{c}{\textbf{ScanNet++}}
& \multicolumn{2}{c}{\textbf{Mean}} \\
\cmidrule(lr){2-3}\cmidrule(lr){4-5}\cmidrule(lr){6-7}\cmidrule(lr){8-9}
& $\mathrm{F1}@0.25$ & $\mathrm{AP}_{3\mathrm{D}}$
& $\mathrm{F1}@0.25$ & $\mathrm{AP}_{3\mathrm{D}}$
& $\mathrm{F1}@0.25$ & $\mathrm{AP}_{3\mathrm{D}}$
& $\mathrm{F1}@0.25$ & $\mathrm{AP}_{3\mathrm{D}}$ \\
\midrule
Qwen3-VL-4B
& 14.3 & 8.4 & 19.2 & 9.3 & 14.5 & 8.1 & 16.0 & 8.6 \\
Qwen3-VL-8B
& 19.4 & 11.2 & 21.2 & 11.2 & 14.4 & 12.4 & 18.3 & 11.6 \\
Qwen3.5-4B
& 10.3 & 7.3 & 27.3 & 11.0 & 8.8 & 8.4 & 15.5 & 8.9 \\
Qwen3.5-9B
& 14.8 & 9.9 & 22.2 & 11.6 & 11.9 & 9.2 & 16.3 & 10.2 \\
VG-LLM
& 27.9 & 11.5 & 25.3 & 18.3 & 22.7 & 13.1 & 25.3 & 14.3 \\
DetAny3D
& 21.0 & 14.4 & 51.3 & 30.6 & 31.3 & 18.5 & 34.5 & 21.2 \\
OVMono3D
& 46.3 & 24.7 & 52.9 & 29.8 & \underline{46.2} & \underline{29.4} & \underline{48.5} & 28.0 \\
\midrule
TempoGround-4B (ours)
& \underline{47.6} & \underline{25.5} & \underline{56.1} & \underline{38.6} & 40.7 & 27.5 & 48.1 & \underline{30.5} \\
TempoGround-9B (ours)
& \textbf{49.5} & \textbf{30.8} & \textbf{63.0} & \textbf{45.3} & \textbf{51.7} & \textbf{30.4} & \textbf{54.7} & \textbf{35.5} \\
\bottomrule
\end{tabular*}
\end{table*}

\begin{table*}[t]
\centering
\small
\setlength{\tabcolsep}{3.2pt}
\caption{$2$D grounding results.
Best results are in \textbf{bold}; second best are \underline{underlined}.}
\label{tab:main-2d}
\begin{tabular*}{\textwidth}{@{\extracolsep{\fill}}lcccccccc@{}}
\toprule
\multirow{2}{*}[-0.5ex]{\textbf{Method}}
& \multicolumn{2}{c}{\textbf{ASE}}
& \multicolumn{2}{c}{\textbf{ARKitScenes}}
& \multicolumn{2}{c}{\textbf{ScanNet++}}
& \multicolumn{2}{c}{\textbf{Mean}} \\
\cmidrule(lr){2-3}\cmidrule(lr){4-5}\cmidrule(lr){6-7}\cmidrule(lr){8-9}
& $\mathrm{F1}@0.5$ & $\mathrm{F1}@0.95$
& $\mathrm{F1}@0.5$ & $\mathrm{F1}@0.95$
& $\mathrm{F1}@0.5$ & $\mathrm{F1}@0.95$
& $\mathrm{F1}@0.5$ & $\mathrm{F1}@0.95$ \\
\midrule
Qwen3-VL-4B
& 36.8 & 5.1 & 36.0 & 11.9 & 54.1 & 19.2 & 42.3 & 12.1 \\
Qwen3-VL-8B
& 38.0 & 2.4 & 34.4 & 11.1 & 54.2 & 17.7 & 42.2 & 10.4 \\
Qwen3.5-4B
& 16.4 & 1.5 & 31.0 & 6.3 & 52.9 & 13.5 & 33.4 & 7.1 \\
Qwen3.5-9B
& 12.1 & 1.7 & 33.7 & 7.1 & 56.4 & 12.7 & 34.1 & 7.2 \\
Cosmos-Reason2-8B
& 12.4 & 2.1 & 33.7 & 6.3 & 43.3 & 7.6 & 29.8 & 5.3 \\
GroundingDINO-T
& \underline{43.4} & \underline{20.1} & 45.7 & 13.8 & \underline{74.5} & \textbf{40.5} & \underline{54.5} & 24.8 \\
GLaMM
& 13.5 & 2.1 & 15.9 & 4.0 & 35.5 & 10.2 & 21.6 & 5.4 \\
RexOmni-3B
& 43.3 & 19.4 & 42.9 & 17.6 & 65.8 & 38.3 & 50.7 & 25.1 \\
LocateAnything
& 35.1 & \textbf{21.7} & \textbf{52.1} & 18.7 & 72.3 & 37.3 & 53.2 & \underline{25.9} \\
\midrule
TempoGround-4B (ours)
& 38.2 & 19.4 & 49.8 & \underline{21.0} & 62.3 & 30.5 & 50.1 & 23.6 \\
TempoGround-9B (ours)
& \textbf{45.5} & 18.6 & \underline{51.9} & \textbf{21.2} & \textbf{79.4} & \underline{39.3} & \textbf{58.9} & \textbf{26.4} \\
\bottomrule
\end{tabular*}
\end{table*}

Table~\ref{tab:main-3d} and Table~\ref{tab:main-2d} report $3$D and $2$D results under causal streaming inputs.
TempoGround consistently surpasses both general-purpose VLMs and specialist grounders, indicating that state-aware curriculum prediction and SGR are effective for online localization rather than only for offline single-frame recognition.
The $2$D gains under the same protocol further show that the benefit is not confined to camera-frame $3$D lifting, but also improves image-plane grounding when predictions must be refreshed causally.

\subsection{Ablation Studies}
\label{sec:ablation}

All ablations keep the same TempoGround-9B backbone, data budget, and evaluation protocol on ScanNet under causal streaming with $\mathrm{AP}_{3\mathrm{D}}$.
Curriculum-prediction ablations use the Stage~2 supervised checkpoint for computational efficiency, while SGR ablations remove SGR or individual rewards from the full model.

\begin{table}[t]
\centering
\small
\setlength{\tabcolsep}{4pt}
\caption{Ablation of curriculum prediction.}
\label{tab:ablation-curriculum}
\begin{tabular*}{\linewidth}{@{\extracolsep{\fill}}l c@{}}
\toprule
\textbf{Configuration} & $\mathrm{AP}_{3\mathrm{D}}$ \\
\midrule
w/o correspondence prior & 37.7 \\
$2$D+$3$D prior & 41.8 \\
w/o presence-state prediction & 40.5 \\
Ours & \textbf{42.3} \\
\bottomrule
\end{tabular*}
\end{table}

\begin{table}[t]
\centering
\small
\setlength{\tabcolsep}{4pt}
\caption{Ablation of SGR rewards.}
\label{tab:ablation-sgr}
\begin{tabular*}{\linewidth}{@{\extracolsep{\fill}}l c@{}}
\toprule
\textbf{Configuration} & $\mathrm{AP}_{3\mathrm{D}}$ \\
\midrule
w/o SGR & 42.3 \\
w/o Grounding reward $R_{\mathrm{g}}$ & 42.7 \\
w/o Identity reward $R_{\mathrm{id}}$ & 44.1 \\
w/o Consistency reward $R_{\mathrm{con}}$ & 43.9 \\
Ours & \textbf{45.3} \\
\bottomrule
\end{tabular*}
\end{table}

Table~\ref{tab:ablation-curriculum} shows that both cross-frame correspondence and presence supervision improve $\mathrm{AP}_{3\mathrm{D}}$, to different degrees.
Carrying camera-frame $3$D boxes in the prior underperforms the $2$D-only cue, consistent with Section~\ref{sec:curriculum}: previous-view $3$D geometry can bias localization under the current camera frame.
Table~\ref{tab:ablation-sgr} shows that Stage~3 SGR brings a clear overall gain, and removing any of Grounding, Identity, or Consistency lowers accuracy, indicating that all three rewards contribute positively.

\subsection{In-depth Analysis}
\label{sec:analysis}

\begin{figure}[!t]
\centering
\includegraphics[width=\linewidth]{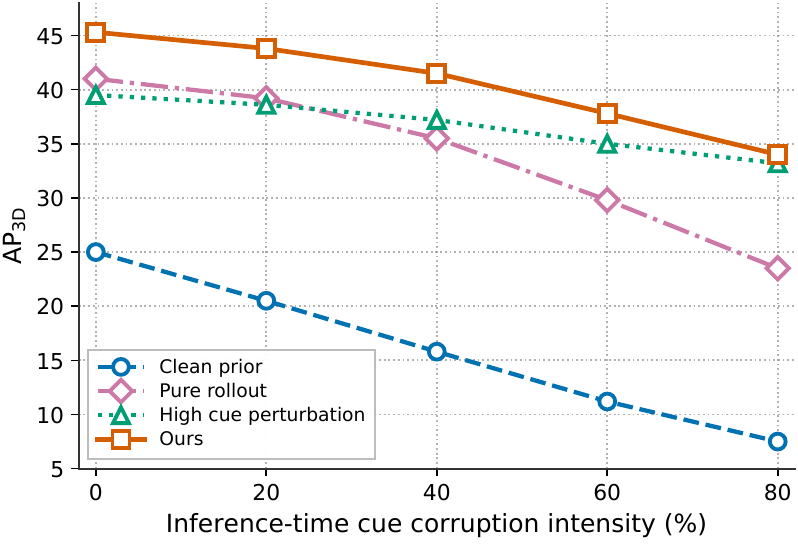}
\caption{Closed-loop robustness to inference-time cue corruption on ScanNet ($\mathrm{AP}_{3\mathrm{D}}$).}
\label{fig:cue-robustness}
\end{figure}

Under closed-loop streaming, errors in $\mathbf{c}_{t-1}$ can propagate across timesteps.
We therefore corrupt the incoming cue at inference with spatial jitter, box dropout, and label mixing, sweeping the corruption intensity from $0\%$ to $80\%$.
Figure~\ref{fig:cue-robustness} compares four Stage~2 cue schedules on ScanNet $\mathrm{AP}_{3\mathrm{D}}$.

Training with clean ground-truth cues alone yields a low closed-loop starting point and a steep decay, revealing an oracle-prior gap.
Pure Stage~1 rollout improves the clean operating point and remains relatively stable under mild corruption, but degrades once the noise exceeds the Stage~1 error distribution.
High cue perturbation flattens the curve yet suppresses peak accuracy, indicating that excessive prior noise weakens useful association signals.
\emph{Ours}, which mixes rollout with the synthetic modes in Table~\ref{tab:cue-perturbation}, attains the best accuracy across the full intensity range: rollout supplies realistic closed-loop residuals, while synthetic perturbations cover rarer failure modes that rollout alone does not.

\section{Conclusion}

We presented TempoGround, a VLM-native framework for accurate and consistent visual grounding under causally arriving image streams.
TempoGround detects cross-frame object correspondence, explicitly models object presence states, and follows a curriculum prediction chain from $2$D association and presence estimation to camera-frame $3$D grounding.
We further introduced Streaming Grounding Reinforcement (SGR), which optimizes TempoGround with verifiable Grounding, Identity, and Consistency rewards beyond token-level imitation.
Extensive experiments demonstrate that TempoGround provides a practical foundation for streaming visual grounding.


\bibliography{aaai2027}


\clearpage
\twocolumn[{%
  \centering
  {\LARGE\bf Supplementary Material\par}
  \vspace{0.4em}
  \vspace{1.2em}
}]
\appendix
\renewcommand{\thefigure}{\thesection.\arabic{figure}}
\renewcommand{\thetable}{\thesection.\arabic{table}}
\renewcommand{\thealgorithm}{\thesection.\arabic{algorithm}}
\makeatletter
\@addtoreset{algorithm}{section}
\@addtoreset{figure}{section}
\@addtoreset{table}{section}
\makeatother

This supplementary material provides details omitted from the main paper.
\textbf{Appendix~\ref{app:impl}} specifies the output schema, the computation of localization metrics, and runtime measurements.
\textbf{Appendix~\ref{app:sgr}} gives the Stage~3 reward formulas used by Streaming Grounding Reinforcement.
\textbf{Appendix~\ref{app:data}} describes data curation: quality control, referring annotation, correspondence-cue perturbation, and prompt templates.
\textbf{Appendix~\ref{app:viz}} shows additional qualitative results.

\section{Implementation Details}
\label{app:impl}

\subsection{Output Schema}
\label{app:output-schema}

TempoGround introduces special tokens that delimit the curriculum prediction fields.
The token inventory is
\begin{center}
\small\ttfamily
\textless ref\textgreater/\textless/ref\textgreater,\ 
\textless state\textgreater/\textless/state\textgreater,\ 
\textless 2d\_bbox\textgreater/\textless/2d\_bbox\textgreater,\ 
\textless 3d\_bbox\textgreater/\textless/3d\_bbox\textgreater,\ 
\textless null\textgreater.
\end{center}
A visible instance with state \texttt{new} or \texttt{track} is serialized as
\begin{quote}\small\ttfamily
\textless ref\textgreater\textit{\rmfamily label}\textless/ref\textgreater\textless state\textgreater\textit{\rmfamily state}\textless/state\textgreater\\
\textless 2d\_bbox\textgreater$x_1,y_1,x_2,y_2$\textless/2d\_bbox\textgreater\\
\textless 3d\_bbox\textgreater[$c_x,c_y,c_z$] [$s_x,s_y,s_z$] [$r,p,y$]\textless/3d\_bbox\textgreater
\end{quote}
where the $2$D coordinates are integers in $[0,1000]$ and the $3$D quantities follow the camera-frame convention of the main paper.
An exited instance is serialized without boxes as
\begin{center}
\small\ttfamily
\textless ref\textgreater\textit{\rmfamily label}\textless/ref\textgreater\textless state\textgreater lost\textless/state\textgreater.
\end{center}
When no query-matching object is visible, the model emits \texttt{<null>}.
Multiple instances are concatenated in a canonical order.
At inference, thinking-mode generation is disabled and the model is served with vLLM.

\subsection{Evaluation Protocol}
\label{app:eval}

We report $\mathrm{F1}_{2\mathrm{D}}@0.5$, $\mathrm{F1}_{2\mathrm{D}}@0.95$, $\mathrm{F1}_{3\mathrm{D}}@0.25$, and $\mathrm{AP}_{3\mathrm{D}}$.
Under the TempoGround protocol, the previous-frame correspondence cue is constructed from the model's own prediction at the preceding timestep.

\paragraph{Scored instances.}
Before scoring, instances with state \texttt{lost} are discarded from both the ground truth and the prediction.
Frames that contain no remaining visible ground-truth box are excluded from $\mathrm{F1}$ and $\mathrm{AP}_{3\mathrm{D}}$.

\paragraph{Matching.}
On each scored frame, correspondence between predictions and ground-truth boxes is established by greedy matching on axis-aligned $2$D IoU.
All candidate pairs are sorted in decreasing order of $2$D IoU, and a pair is accepted only when neither side has already been matched.
The resulting one-to-one assignment is shared by $2$D and $3$D $\mathrm{F1}$: correspondence is determined by $2$D overlap, after which oriented $3$D IoU is evaluated on the matched pairs.

\paragraph{F1 score.}
Let $N_t^{\mathrm{pred}}$ and $N_t^{\mathrm{gt}}$ denote the numbers of predicted and visible ground-truth boxes on a scored frame $t$, and let $\mathrm{TP}_t(\delta)$ denote the number of matched pairs whose overlap is at least $\delta$.
For $\mathrm{F1}_{2\mathrm{D}}@\delta$ the overlap is $2$D IoU; for $\mathrm{F1}_{3\mathrm{D}}@\delta$ it is oriented $3$D IoU on the same pairs.
Precision and recall are
\begin{equation}
P_t(\delta)=\frac{\mathrm{TP}_t(\delta)}{N_t^{\mathrm{pred}}},
\qquad
R_t(\delta)=\frac{\mathrm{TP}_t(\delta)}{N_t^{\mathrm{gt}}},
\end{equation}
with the convention that a zero denominator yields zero.
The per-frame score is
\begin{equation}
\mathrm{F1}_t(\delta)=\frac{2\,P_t(\delta)\,R_t(\delta)}{P_t(\delta)+R_t(\delta)},
\end{equation}
or $0$ when $P_t(\delta)+R_t(\delta)=0$.
Within each dataset we take the mean of $\mathrm{F1}_t$ over scored frames; the reported number is then the unweighted mean over datasets.

\paragraph{Average precision.}
$\mathrm{AP}_{3\mathrm{D}}$ pools predictions and ground-truth boxes from all scored frames in a dataset.
At each oriented $3$D IoU threshold $\delta\in\Theta=\{0.15,\,0.25,\,0.50\}$, predictions are matched greedily to unused ground-truth boxes, and AP is computed with the Pascal VOC $11$-point interpolation.
Detection averages AP over categories; referring uses a single referred target.
The dataset-level score is
\begin{equation}
\mathrm{AP}_{3\mathrm{D}}=\frac{1}{|\Theta|}\sum_{\delta\in\Theta}\mathrm{AP}(\delta),
\end{equation}
and the reported number is the unweighted mean over datasets.

\subsection{Runtime}
\label{app:runtime}

As shown in Table~\ref{tab:runtime}, the average single-image inference latency of TempoGround on one NVIDIA H200 GPU is $0.65$\,s for the $9$B model and $0.45$\,s for the $4$B model, each averaged over $10$ runs.
The reported time covers the full forward path for one observation, including vision encoding of the input image, prefilling, and autoregressive generation of the grounding outputs.
These results show that TempoGround enables a general-purpose VLM to perform continuous object grounding at sub-second speed, providing a practical foundation for general embodied perception and downstream tasks such as spatial reasoning and question answering.

\begin{table}[h]
\centering
\begin{tabular*}{\columnwidth}{@{\extracolsep{\fill}}lc@{}}
\toprule
Model & Mean Latency (s) \\
\midrule
TempoGround-9B & $0.65$ \\
TempoGround-4B & $0.45$ \\
\bottomrule
\end{tabular*}
\caption{Average inference latency on one NVIDIA H200 GPU over $10$ runs.}
\label{tab:runtime}
\end{table}

\section{Streaming Grounding Reinforcement}
\label{app:sgr}

The main paper defines the Stage~3 objective and the roles of $R_{\mathrm{g}}$, $R_{\mathrm{id}}$, and $R_{\mathrm{con}}$.
This appendix specifies how each term is computed.

\paragraph{Shared geometric quality.}
After pairing a visible ground-truth instance with a prediction, we define
\begin{equation}
q
=
0.35\,\mathrm{IoU}_{2\mathrm{D}}
+
0.65\,\mathrm{IoU}_{3\mathrm{D}}.
\end{equation}
If the instance has no match, $q{=}0$.
This scalar is reused by $R_{\mathrm{g}}$ and $R_{\mathrm{id}}$.

\subsection{Grounding Reward $R_{\mathrm{g}}$}
\label{app:sgr-grounding}

$R_{\mathrm{g}}$ measures trajectory-level localization quality while penalizing unmatched predictions.
It is not a single global average of all per-frame $q$ values.
A global average would let long tracks dominate short ones and would ignore spurious boxes that never match any ground-truth instance.
The implementation therefore separates localization credit from false-positive cost.

For each ground-truth track $i$, let $\mathcal{T}_i$ be the set of frames on which its annotated presence is $\texttt{new}$ or $\texttt{track}$.
The track score is the mean quality on those frames,
\begin{equation}
r_i
=
\frac{1}{|\mathcal{T}_i|}\sum_{t\in\mathcal{T}_i} q_{i,t}.
\end{equation}
Equal weight is then given to every track,
\begin{equation}
R_{\mathrm{track}}
=
\frac{1}{M}\sum_{i=1}^{M} r_i,
\end{equation}
where $M$ is the number of tracks.
Frames annotated as $\texttt{lost}$ are excluded from $r_i$ because no box is required there; presence errors on those frames are handled by $R_{\mathrm{id}}$.

$R_{\mathrm{track}}$ alone is insufficient: a policy can keep $R_{\mathrm{track}}$ high while emitting many unmatched boxes.
Let $N_{\mathrm{fp}}$ be the number of unmatched predictions over the trajectory and $N_{\mathrm{GT}}$ the number of ground-truth instances over the trajectory.
The grounding reward subtracts a normalized false-positive penalty and clips to $[0,1]$:
\begin{equation}
R_{\mathrm{g}}
=
\mathrm{clip}\!\left(
R_{\mathrm{track}}
-
0.3\,\frac{N_{\mathrm{fp}}}{\max(1,N_{\mathrm{GT}})},
\,0,1\right).
\end{equation}
Normalization by $N_{\mathrm{GT}}$ keeps the penalty comparable across trajectories of different length and object count.

\subsection{Identity Reward $R_{\mathrm{id}}$}
\label{app:sgr-identity}

$R_{\mathrm{id}}$ is the mean of per-instance event scores over the trajectory, clipped to $[-1,1]$.
Each ground-truth instance on each frame contributes one event score according to its annotated presence:

\begin{itemize}
\item $\texttt{new}$:
matched $\texttt{new}$ gives $+q$;
matched $\texttt{track}$ gives $-0.5$;
a miss or any other predicted state gives $-1$.
\item $\texttt{track}$:
matched $\texttt{track}$ gives $0$;
a miss or predicted $\texttt{lost}$ gives $-\alpha$, where $\alpha\in[0,1]$ is the visible fraction of the object's $2$D box after clipping to the image boundary;
any other incorrect predicted state gives $-0.3$.
\item $\texttt{lost}$:
a matched non-$\texttt{lost}$ prediction gives $-1$;
otherwise the event gives $0$.
\end{itemize}

\subsection{Consistency Reward $R_{\mathrm{con}}$}
\label{app:sgr-consistency}

The correspondence cue at timestep $t{+}1$ is constructed from the prediction at timestep $t$.
A localization error is therefore not confined to the current frame: it can corrupt later association and produce a contiguous run of unstable updates.
Adjacent-frame flicker is a visible symptom of this process, but penalizing differences between consecutive predictions is ill-posed under ego-motion, where boxes are expected to change.
$R_{\mathrm{con}}$ instead checks each frame against ground truth and penalizes failures that form long contiguous chains, namely the self-conditioned streaks that cue feedback amplifies.

\paragraph{Frame failure.}
After $2$D greedy matching of visible ground truth, a frame is marked as a localization failure if either
(i)~at least one prediction remains unmatched, or
(ii)~the mean quality $\bar q$ over visible ground-truth instances falls below $0.25$.
A frame with no visible ground truth and no prediction is not a failure.

\paragraph{Phantom after exit.}
A frame is marked as a phantom frame when the ground truth contains at least one $\texttt{lost}$ instance and unmatched predictions remain after the still-visible targets have been matched.
If all query objects have left and predictions remain nonempty, the frame is also marked as a phantom frame.
This criterion targets continued emission after exit without penalizing boxes on objects that remain visible.

\paragraph{Chain penalty.}
Let $\{m_t\}_{t=1}^{T}$ and $\{p_t\}_{t=1}^{T}$ be the binary failure and phantom indicators along the trajectory.
For a binary sequence, let $\ell_1,\ell_2,\ldots$ be the lengths of its maximal contiguous runs of ones.
The chain cost is
\begin{equation}
\mathrm{chain}(\{z_t\})
=
\frac{1}{T^2}\sum_k \ell_k^{2}.
\end{equation}
An isolated error contributes little; a long unbroken run grows with the square of its length.
The consistency reward is
\begin{equation}
R_{\mathrm{con}}
=
-\frac12\Big(
\mathrm{chain}(\{m_t\})
+
\mathrm{chain}(\{p_t\})
\Big),
\end{equation}
clipped to $[-1,0]$.
IoU decides whether the current step is wrong; $R_{\mathrm{con}}$ decides whether those wrongs have turned into a self-reinforced streak under cue feedback.

In Stage~3 training, $R_{\mathrm{g}}$, $R_{\mathrm{id}}$, and $R_{\mathrm{con}}$ are mixed with weights $0.70$, $0.15$, and $0.15$.
Let $N_{\mathrm{empty}}$ be the number of frames whose prediction text is empty while the ground truth is nonempty, and define
\begin{equation}
f_{\mathrm{fmt}}
=
\frac{N_{\mathrm{empty}}}{T}.
\end{equation}
The scalar trajectory reward subtracts $0.02\,f_{\mathrm{fmt}}$.
Empty frames still enter the geometric terms with no parsed boxes; the format term is an additional penalty on the empty-output rate.

\section{Data Curation Details}
\label{app:data}

\subsection{Quality Control}
\label{app:data-qc}

Beyond the high-level frustum and visibility filters in the main paper, the streaming builders use the following two geometric gates, transition rules, and sequence acceptance criteria.

\paragraph{In-view versus present.}
Let $A_{\mathrm{amodal}}$ be the area of the projected $2$D box before image clipping, $A_{\mathrm{clip}}$ the area after clipping to the image, and $A_{\mathrm{img}}$ the image area.
An instance is \emph{in view} if its camera-frame depth is positive ($z>0.1$), $A_{\mathrm{clip}}\ge 64$ pixels, and the clipped fraction $A_{\mathrm{clip}}/A_{\mathrm{amodal}}$ is at least $0.05$.
This gate only asks whether the projection still intersects the camera frustum with nontrivial mass.

Positive localization supervision further requires the stricter \emph{present} gate.
Present demands $A_{\mathrm{clip}}/A_{\mathrm{amodal}}\ge 0.30$ and $A_{\mathrm{clip}}/A_{\mathrm{img}}\ge 2\times 10^{-4}$.
When pixel-level visibility is available, the visible-pixel fraction inside the box must be at least $0.15$; if the clipped box covers at least $25\%$ of the image, the ratio of visible pixels to box-image fraction must also be at least $0.20$, so a large empty amodal box is rejected.
When only mesh-extent visibility is available, we require extent fraction at least $0.35$ (with a weak band down to $0.30$).
Boxes that are in view but fail present may still participate in enter/exit bookkeeping; they do not contribute positive $2$D/$3$D regression targets.

\paragraph{Presence-state transitions.}
States are assigned from the in-view sequence, not from raw preprocessing files.
For detection, an object is supervisable when it is in view with a box; referring adds a dependency check described in Appendix~\ref{app:data-refer}.
An object becomes $\texttt{new}$ when it first becomes supervisable in a track, or when it re-enters after exit.
While it remains in view after that enter event it is $\texttt{track}$.
The $\texttt{lost}$ label is emitted on exactly one frame: the transition from in-view to out-of-view.
That frame carries no $2$D or $3$D box and is never written into the next-step correspondence cue.
Subsequent out-of-view frames are empty responses rather than repeated $\texttt{lost}$ tokens with a stale box.
Re-entry again uses $\texttt{new}$.
Automated track assertions reject lost streaks longer than one frame, boxes on $\texttt{lost}$ or empty frames, and illegal $\texttt{new}$/$\texttt{track}$ transitions.

\paragraph{Sequence acceptance and source hygiene.}
A retained streaming window must contain an enter event and either an exit event or a leading empty prefix, so that the clip is not a static always-centered fragment.
Along the window of length $T$, the number of present frames must be at least $\max(6,\lfloor 0.4T\rfloor)$, and the best present frame must reach visible-pixel fraction $0.20$ (or extent $0.40$ when pixels are unavailable).
Training windows use lengths in $[24,64]$ and test windows in $[32,64]$, with dataset-dependent base strides, multiplicative jitter on ScanNet-family sources, and a small integer offset so neighboring clips are not identical.
Windows that prefer enter/persist/exit transitions are favored over near-static runs.
Category labels are canonicalized across sources; large structural or geometrically ill-posed classes (for example wall, floor, and ceiling) are removed or heavily down-weighted.
For ScanNet++, frames whose color field is nearly flat are rejected as appearance-unstable.
Calibrated negatives are retained so that a query with no matching in-view object supervises abstention.

\subsection{Referring Expression Annotation}
\label{app:data-refer}

Public scene-level expressions are discarded as primary supervision, as stated in the main paper.
We detail the annotation inputs, frame-selection thresholds, programmatic gates, and uniqueness checks used when regenerating view-grounded queries.

\paragraph{Index-only use of public refer corpora.}
ScanRefer, Nr3D, and related releases contribute only object indices for the referring domain.
All pixels and camera-frame boxes come from the underlying RGB-D scans (for example ScanNet), not from the scene-level expression text or map-centric wording in those corpora.

\paragraph{Annotation-frame selection.}
A frame is eligible for writing an expression only if the target is sufficiently complete and the view retains surrounding context.
Concretely, the clipped-to-amodal completeness must be at least $0.90$, the visible-pixel fraction at least $0.35$ when available, and the target box must cover less than $40\%$ of the image while leaving at least $25\%$ of the image as non-target context.
At least one co-visible neighbor that itself meets milder completeness thresholds must remain in the allowlist; otherwise the frame is discarded because relational language has no verified anchor.
When multiple views of the same object are kept, each view is represented by a layout signature that quantizes the target box onto a $10\times 10$ grid and records the coarse image regions of co-visible neighbors.
Signatures that differ by at most one grid cell with an identical neighbor set are treated as near-duplicates and are not both sent to the annotator.

\paragraph{Proposal and programmatic gates.}
On the selected primary frame, the target is rendered with a salient green box; neighbors appear only as a text allowlist that lists official category names, object indices, and coarse image regions.
A closed-source vision-language model (Qwen3.7-Plus) proposes three short candidate expressions per target, preferring intrinsic attributes and using allowlisted neighbors only when attributes cannot separate same-category distractors.
Candidates are discarded if they mention doorway- or entrance-centric viewpoint language, cardinal directions, hedged uncertainty phrases, or any dependency outside the allowlist.
When more than one same-category instance is co-visible, left/right or foreground/background alone is also forbidden as the sole discriminator, and at least one allowlisted neighbor dependency is required.

\paragraph{Uniqueness verification.}
If several same-category instances remain co-visible, two visual checks are applied before the expression is accepted.
In the candidate-index check, every same-category box is numbered on the image and the model must return the target index given only the candidate sentence.
In the contrastive check, the target is marked in green and one distractor in red; the model must select the green box for that sentence.
Failure of either check discards the paraphrase.
Accepted expressions are attached to the object track for streaming packaging.

\paragraph{Streaming resolve-then-track.}
Along a referring track, a frame is labeled $\texttt{new}$ only when the target is in view and every linguistic dependency in the expression is co-visible (the query is uniquely decidable from the observed view).
After that enter event, subsequent $\texttt{track}$ labels follow in-view geometry under cross-frame correspondence even if some anchors leave the frame.
Exit uses the same one-frame box-free $\texttt{lost}$ transition as detection; re-entry again requires the dependencies to be co-visible.
Frames that are geometrically visible but not yet decidable remain empty rather than receiving an under-specified $\texttt{new}$.

\subsection{Correspondence-Cue Perturbation}
\label{app:cue-perturbation}

Stage~2 constructs the correspondence cue $\mathbf{c}_{t-1}$ while keeping the supervised target at timestep $t$ unchanged.
The cue contains only instance labels and previous-frame $2$D boxes; camera-frame $3$D is never carried forward.
For each streaming sample a mode is sampled according to the mixture in the main paper, and the selected operator is applied to the previous-frame ground-truth cue (or replaces it entirely).
We detail the operators below.

\paragraph{Clean.}
$\mathbf{c}_{t-1}$ is the set of ground-truth labels and $2$D boxes on frame $t{-}1$.

\paragraph{Stage~1 rollout.}
A frozen Stage~1 checkpoint is run on frame $t{-}1$ under the same query.
Its predicted labels and $2$D boxes replace the ground-truth cue, so $\mathbf{c}_{t-1}$ follows the closed-loop residual distribution of the Stage~1 policy rather than an oracle prior.

\paragraph{Jitter.}
Starting from the clean previous-frame cue, each $2$D box is translated independently.
For every instance a magnitude $m$ is drawn uniformly from the integer range $[5,30]$ in the normalized coordinate scale $[0,1000]$, after which horizontal and vertical offsets are sampled uniformly from $[-m,m]$ and mapped to pixels by the image width and height.
All four corners of that box receive the same offset, so shape is preserved and only the association location is displaced.

\paragraph{Dropout.}
Starting from the clean cue, a random nonempty subset of prior instances is removed.
If the clean cue contains $N$ instances, the number of removals is drawn uniformly from $1$ to $\max(1,\lfloor N/2\rfloor)$, after which the remaining instances are kept as $\mathbf{c}_{t-1}$.

\paragraph{Phantom.}
One or two spurious instances are appended to the clean cue.
Each phantom copies an existing prior box when available, perturbs every coordinate by an independent offset up to $20\%$ of the image width or height, and replaces the label by a category sampled from the dataset vocabulary when that vocabulary is available.
If the clean cue is empty, phantoms are instead sampled as random boxes in the image with vocabulary labels.

\paragraph{Stale.}
Instead of the cue from frame $t{-}1$, $\mathbf{c}_{t-1}$ is taken from an older frame in the same stream, either $t{-}2$ or $t{-}3$ when available.
The older cue is otherwise left uncorrupted, so the association signal is temporally outdated rather than spatially jittered.

\paragraph{Empty.}
$\mathbf{c}_{t-1}$ is set to the empty set and rendered as a null prior in the prompt.
This matches a stream restart: the current frame must be resolved without cross-frame correspondence.

\subsection{Prompt Templates}
\label{app:prompts}

Streaming prompts follow the curriculum prediction interface: presence, then $2$D, then camera-frame $3$D.
First-frame prompts omit the previous image and cue; streaming prompts include both.
Detection prompts request every instance of a category; referring prompts request a single expression.
Each template family has a small set of paraphrases selected deterministically from the sample key, so surface wording varies without changing the output schema.

A first-frame detection prompt is
\begin{quote}
\ttfamily\footnotesize
Current frame: <image>\\
Detect every ``category'' in the current frame. For each visible instance, output a 2D box then a 3D bounding box in the camera frame.
\end{quote}
A streaming detection prompt is
\begin{quote}
\ttfamily\footnotesize
Previous frame: <image>\\
Current frame: <image>\\
Previous-frame detections for ``category'' (label with 2D box): prior\\
Using the previous frame as reference, detect every ``category'' in the current frame. For each object, report its status (newly appeared, still tracked, or gone) and output a 2D box then a 3D box in the camera frame.
\end{quote}
Referring prompts use the same first-frame and streaming split, with the category query replaced by the regenerated expression and a single referred target.
Prompted status terms are aligned with the supervised presence vocabulary, and geometric fields follow Appendix~\ref{app:output-schema}.

\section{Additional Qualitative Results}
\label{app:viz}

\begin{figure*}[t]
\centering
\includegraphics[width=0.6\textwidth]{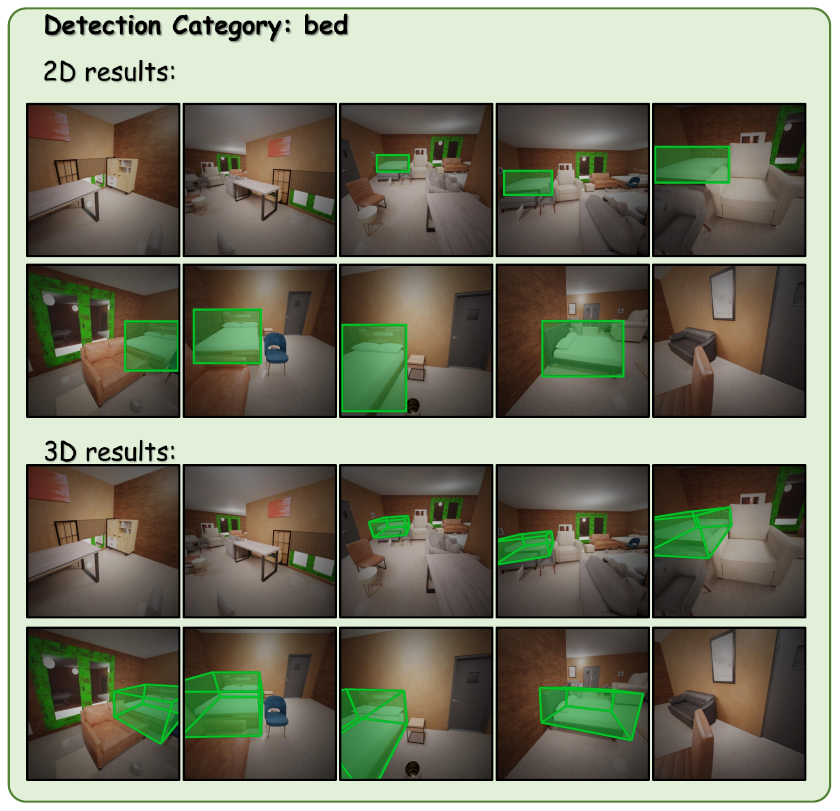}
\caption{While the bed remains in the field of view, TempoGround maintains correct $2$D/$3$D tracking.}
\label{fig:supp-det-bed}
\end{figure*}

\begin{figure*}[t]
\centering
\includegraphics[width=0.6\textwidth]{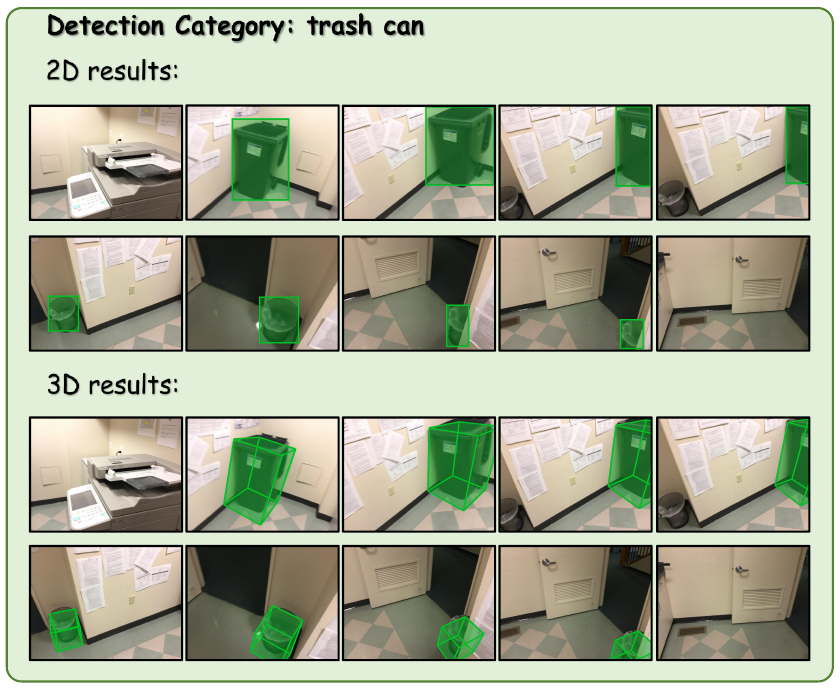}
\caption{The scene contains both a large and a small trash can; TempoGround accurately detects the two instances.}
\label{fig:supp-det-trashcan}
\end{figure*}

\begin{figure*}[t]
\centering
\includegraphics[width=0.6\textwidth]{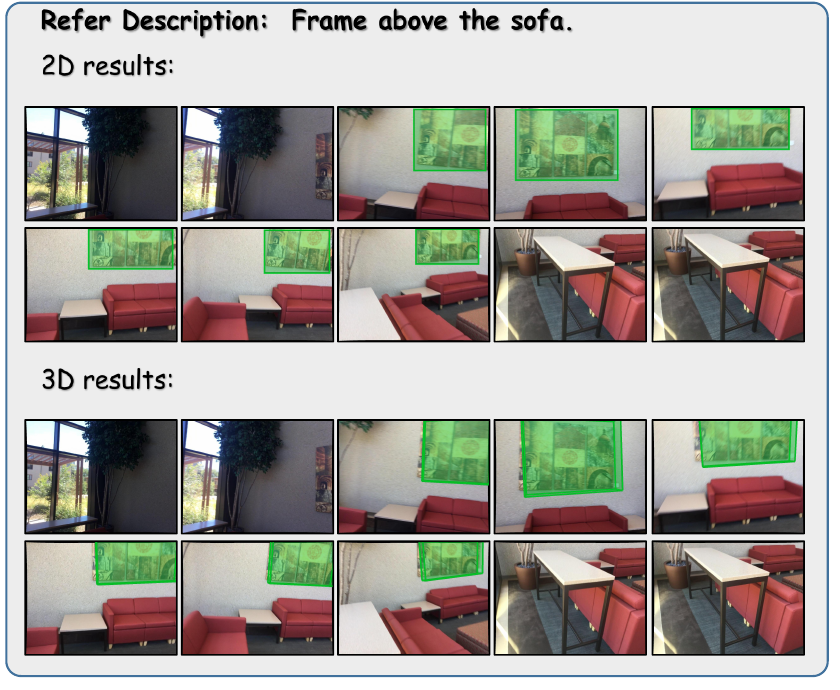}
\caption{When the frame above the sofa enters the view, TempoGround tracks it accurately in $2$D and $3$D.}
\label{fig:supp-refer-frame}
\end{figure*}

\begin{figure*}[t]
\centering
\includegraphics[width=0.6\textwidth]{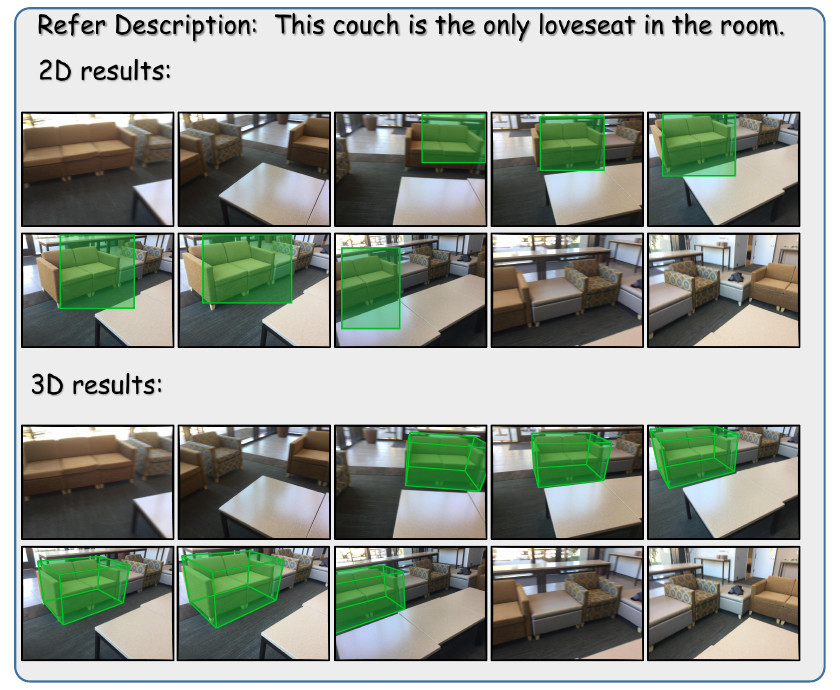}
\caption{TempoGround interprets the query ``loveseat'' and correctly identifies the two-seat couch in the view, avoiding false detections of single-seat or three-seat ones.}
\label{fig:supp-refer-loveseat}
\end{figure*}

We show additional qualitative visualizations for streaming detection (Figures~\ref{fig:supp-det-bed} and~\ref{fig:supp-det-trashcan}) and referring (Figures~\ref{fig:supp-refer-frame} and~\ref{fig:supp-refer-loveseat}).

\end{document}